\documentclass{article}

     \usepackage[preprint]{neurips_2019}

\usepackage[utf8]{inputenc} % allow utf-8 input
\usepackage[T1]{fontenc}    % use 8-bit T1 fonts
\usepackage{hyperref}       % hyperlinks
\usepackage{url}            % simple URL typesetting
\usepackage{booktabs}       % professional-quality tables
\usepackage{amsfonts}       % blackboard math symbols
\usepackage{nicefrac}       % compact symbols for 1/2, etc.
\usepackage{microtype}      % microtypography
\usepackage{amsmath}
\usepackage{multirow}
\usepackage{graphicx}
\graphicspath{{figures/}}

\newcommand{\ie}{\textit{i}.\textit{e}.}

\title{Regularizing Semi-supervised Graph Convolutional Networks with a Manifold Smoothness Loss}

\author{%
  Qilin Li\thanks{Corresponding author}, \quad Wanquan Liu, \quad Ling Li \\
  Discipline of Computing\\
  Curtin University\\
  Perth, Australia \\
  \texttt{li.qilin@postgrad.curtin.edu.au} \\
}

\begin{document}

\maketitle

\begin{abstract}
Existing graph convolutional networks focus on the neighborhood aggregation scheme. When applied to semi-supervised learning, they often suffer from the overfitting problem as the networks are trained with the cross-entropy loss on a small potion of labeled data. In this paper, we propose an unsupervised manifold smoothness loss defined with respect to the graph structure, which can be added to the loss function as a regularization. We draw connections between the proposed loss with an iterative diffusion process, and show that minimizing the loss is equivalent to aggregate neighbor predictions with infinite layers. We conduct experiments on multi-layer perceptron and existing graph networks, and demonstrate that adding the proposed loss can improve the performance consistently.  
\end{abstract}

\section{Introduction}
The success of machine learning methods heavily depends on having good data representation (or features) on which they are applied. Due to this reason, a large portion of efforts in applying machine learning algorithms has been devoted to the design of data preprocessing, transformation, and feature extraction that output a sensible representation of the data so as to support effective machine learning. Conventionally, this is achieved by a labor-intensive feature engineering process that takes advantage of human ingenuity and prior knowledge to produce hand-crafted feature descriptors, such as LBP, SIFT, and HOG used in computer vision. Despite its effectiveness of representing data, hand-crafted features require expensive efforts of highly skilled experts, and the acquired feature descriptors usually do not adapt to different problems, which hinders the efficient deployment of machine learning systems in various real-world applications.

Deep learning, on the other hand, makes use of the depth as the prior knowledge and constructs a deep architecture to extract 
layer-wise representations of the data. More importantly, the layer-wise structure of representation and downstream predictive tasks, such as classifications, can be composed together to a neural network so that they can be trained simultaneously in an end-to-end manner. Features are not anymore crafted by a hand-engineered process but learned automatically via a data-driven and task-driven approach. Feature extraction is not viewed as a preprocessing step of machine learning but treated as the machine learning task itself. It is sometimes referred to as representation learning, one of the most active research areas in the field of machine learning.

Deep learning techniques not only significantly reduce the human efforts required in feature engineering, but also bring breakthroughs in terms of the performance of machine learning algorithms on various applications, such as speech recognition \cite{hinton2012deep}, object classification \cite{krizhevsky2012imagenet}, and natural language processing~\cite{devlin2018bert}. We now have machine learning systems that can recognize images with better accuracies than that of human, drive cars automatically in the real-world environment, and beat human professional players of the board game Go on a full-sized 19$\times$19 board, which is claimed to be the most difficult game in the world. 

While deep learning techniques continue on making progress in various fields, researchers start to think about the limitation of these approaches. They find that classic deep learning techniques are designed for dealing with structured data, such as grids or sequences. For example, convolutional neural networks (CNNs), which is the state-of-the-art technique for all visual data related tasks, can only work with images/grids. These images are grid-style data in which the spatial locality and node ordering are fixed so that the convolutional operation can be directly applied. Recurrent neural networks (RNNs), which is another popular architecture widely used in NLP, can only take texts/sequences as input. These state-of-the-art architectures are powerful as they take advantage of the structured input, but it also limits the application to other non-trivial data structures, such as graphs.

\begin{figure}[htb]
\begin{center}
  \includegraphics[width=0.9\linewidth]{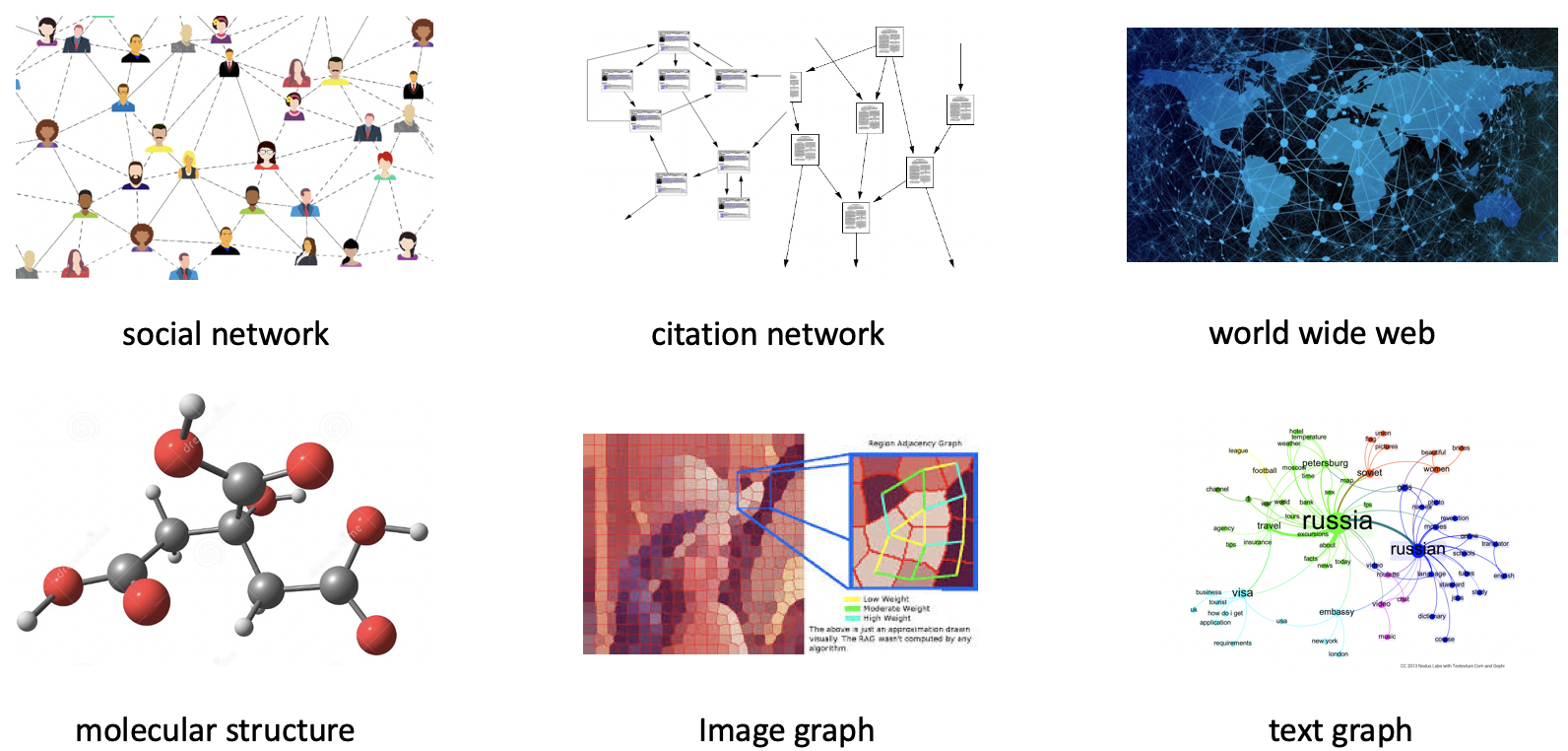}
\end{center}
  \caption{Examples of graph structure}
  \label{ch7:fig:graphs}
\end{figure}

Graphs are a ubiquitous data structure, which is widely employed in computer science, biology, medical science, and many other fields. As a generic topological structure, graphs exist in various forms. Social networks, world wide web, citation networks, molecular structure, biological protein-protein networks are all readily modeled as graphs, including images and texts, as shown in Fig.\ref{ch7:fig:graphs}. In conventional machine learning, graphs are not only useful as a data structure, but also play a critical role in numerous learning models, including spectral graph theory \cite{chung1997spectral}, probabilistic graphical models\cite{koller2009probabilistic}, \emph{etc}. Despite the huge potential of replicating the success of deep learning on graphical data, it does not draw much attention until recent years.

The concept of graph neural networks (GNNs) was first proposed in \cite{scarselli2008graph}, where a basic neural network framework that works on graphical data was created. Instead of using fully connected neural networks, there is a growing interest in applying convolutional operations on graphs, due to the great success of CNN in computer vision. However, it is not straightforward to do so, as nodes on graphs have no obvious order or fixed neighboring structure so that a convolutional filter cannot be readily defined. \cite{hammond2011wavelets} attempted to implement a spectral convolution in the Fourier domain, where an eigenvector problem needs to be solved for the graph Laplacian matrix. \cite{defferrard2016convolutional} extended this idea with a localized constraint that forces neighbors to be within $k$ hops and constructed a CNN on graphs. \cite{kipf2017semi} further extended the spectral graph convolutions with a first-order approximation, resulting in a clear layer-wise propagation formula that can be used to stack a multi-layer architecture similar to CNNs. They named their method as graph convolutional networks (GCNs) and demonstrated superior performances on semi-supervised classification. Since then, a large body of research has been presented in this subfield, including GraphSAGE \cite{hamilton2017inductive}, GAT \cite{velivckovic2017graph}, APPNP \cite{klicpera2018predict}, to name a few.

These approaches generally focus on learning a discriminative representation so that classification can be easily achieved in the feature space. The essential idea of GCNs is to learn the representation of a target node by aggregating the representation of its neighboring nodes. In doing so, the topological graph information is encoded in the local neighborhood structure and utilized in a convolutional way. Neighbor aggregation, linear transformation, and non-linear mapping are integrated as a single layer operation. A deep GCN can be constructed by stacking multiple layers, and it can be trained with backpropagation by calculating a loss with respect to a particular objective function. For example, in semi-supervised classification, the cross-entropy loss on labeled points are often used, while unlabeled points are only used for modeling the graph structure. The limitations are two folds: 1) overfitting. The model is not well trained and does not generalize well due to the lack of training data. This means we cannot use a deep architecture which is proven to be useful for learning good representations \cite{he2016deep}, as it makes the problem of overfitting even worse. This brings another problem: 2) the capacity of modeling long-range neighbor relationships. Existing GCNs usually contains two or three layers, which means they can only aggregate neighbor information as far as 2 or 3 hops away. This significantly limits the model capacity of capturing the global graph structure.

In this work, we propose a novel loss function contains two terms, namely the fitness term and smoothness term. While the fitness term measures the difference between the prediction and ground-truth, the smoothness term regularizes the prediction to be smooth on the underlying manifold represented by the graph. By doing so, the semi-supervised classification can be viewed as a function estimation problem that favors a smooth function with fixed values at certain points. The key point here is that the smoothness term is an unsupervised loss that makes use of both labeled and unlabeled points, which prevents overfitting to the labeled points by the supervised fitness loss only. In addition, we show that the proposed objective function is equivalent to an iterative diffusion process that aggregates neighbor information similar to the layer-wise operation of GCN. Optimizing the objective function is somewhat like going through a GCN with potentially infinite layers, which increases the model capacity significantly. We demonstrate the effectiveness of the proposed loss with semi-supervised classification, where we show that the performance of the standard multi-layer perceptron (MLP) can be significantly boosted when used with the smoothness loss, and also it is ready to be plugged in any state-of-the-art GCNs to improve their performances as well.

\section{Related Work}
We start by reviewing several state-of-the-art GCNs and show that the main difference is on the way of aggregating information of neighbors. Let $\mathcal{G}=(\mathcal{V},\mathcal{E})$ be an undirected graph containing a node set $\mathcal{V}$, an edge set $\mathcal{E}$. Every node has an input feature representation $\mathbf{x}_v \in \mathbb{R}^{d_i}$ for $v \in \mathcal{V}$, and its hidden feature representation learned by $l$-th layer is denoted by $\mathbf{h}_v^{(l)} \in \mathbb{R}^{d_l}$. We may use $\mathbf{h}_v^{(0)} = \mathbf{x}_v$ as the input representation to simplify derivations. Every node contains a neighbor set $\mathcal{N}(v)=\{u \in \mathcal{V} | (v,u) \in \mathcal{E}\}$ that indicates its adjacency nodes. It is common to add the node itself to its neighbor set in GCNs, and we denote it as $\widetilde{\mathcal{N}}(v)=\{u \in \mathcal{V} | (v,u) \in \mathcal{E}\} \cup \{v\}$.

a typical graph convolution layer can be decomposed into three components, including neighbor aggregation, linear transformation, and non-linear mapping. In general, the $l$-th layer of a GCN model updates the hidden representation $\mathbf{h}_v^{(l)}$ as:

\begin{equation}
\mathbf{h}_v^{(l)} = \sigma \, \Bigg(\mathbf{W}^{(l)} \cdot \text{AGGREGATE}\left(\{\mathbf{h}_u^{(l-1)}, \ \forall u \in \widetilde{\mathcal{N}}(v)\}\right)\Bigg),
\end{equation}
where AGGREGATE is an aggregation function defined by the specific model, $\mathbf{W}^{(l)}$ denotes a trainable weight matrix of $l$-th layer shared by all the nodes, and $\sigma$ is a non-linear activation function, such as ReLU.

\textbf{Graph Convolutional Networks (GCN)} \cite{kipf2017semi} is an instance of the general framework with an additional normalization trick, which can be written in the form:

\begin{equation}
\mathbf{h}_v^{(l)} = \sigma \, \left(\mathbf{W}^{(l)} \sum_{u \in \widetilde{\mathcal{N}}(v)} \frac{\mathbf{h}_u^{(l-1)}}{\sqrt{|\mathcal{N}(v)||\mathcal{N}(u)|}}\right),
\end{equation}
where $|\mathcal{N}(v)|$ is the cardinality of the neighbor set $\mathcal{N}(v)$, \ie, the number of neighbors of node $v$. GCN imposes a node-wise normalization in order to balance the effect of each node. The aggregation scheme of GCN can be viewed as a weighted average of all neighbors, and the weighting is proportional to the inverse of the number of neighbors. The update scheme of GCN can be efficiently implemented using sparse batch operations:

\begin{equation}
\mathbf{H}^{(l)} = \sigma \left(\mathbf{D}^{-\frac{1}{2}} \widetilde{\mathbf{A}} \mathbf{D}^{-\frac{1}{2}} \mathbf{H}^{(l-1)} \mathbf{W}^{(l)}\right),
\end{equation}
where $\widetilde{\mathbf{A}} = \mathbf{A} + \mathbf{I}$ is adjacency matrix with self-loop, $\mathbf{D}$ is the diagonal degree matrix with $D_{ii}=\sum_j \widetilde{A}_{ij}$. Indeed, $\mathbf{D}^{-\frac{1}{2}} \widetilde{\mathbf{A}} \mathbf{D}^{-\frac{1}{2}}$ is the normalized graph Laplacian, and actually GCN has a close relation with spectral graph theory \cite{kipf2017semi}.

\textbf{Graph Attention Networks (GAN)} \cite{velivckovic2017graph} can be seen as a variant of GCN with trainable aggregation weights. It utilizes the attention mechanism \cite{bahdanau2014neural} to learn pairwise weights so as to select important neighbors. The update scheme of a single layer in GAN can be written as:

\begin{equation}
\mathbf{h}_v^{(l)} = \sigma \, \left(\sum_{u \in \widetilde{\mathcal{N}}(v)} \alpha_{vu} \mathbf{W}^{(l)} \mathbf{h}_u^{(l-1)} \right),
\end{equation}
where $\alpha_{vu}$ denotes the trainable attention weights between node $v$ and $u$. In GAN, the attention mechanism is implemented as a single-layer feed-forward neural network that can be represented as follows:

\begin{equation}
\alpha_{vu} = \frac{\exp \left(\text{LeakyReLU}\Big(\mathbf{a}^\top[\mathbf{W}\mathbf{h}_v||\mathbf{W}\mathbf{h}_u]\Big) \right)}{\sum_{u \in \widetilde{\mathcal{N}}(v)} \exp \left(\text{LeakyReLU}\Big(\mathbf{a}^\top[\mathbf{W}\mathbf{h}_v||\mathbf{W}\mathbf{h}_u]\Big) \right)},
\end{equation}
where $\mathbf{a}$, $\mathbf{W}$ are the trainable weights and $||$ is the concatenation operation. 

\textbf{Approximate Personalized Propagation of Neural Predictions (APPNP)} \cite{klicpera2018predict} decomposes neighbor aggregation and layer-wise propagation into two steps. It applies the aggregation as a post-processing step after representation learning, which aggregates the neighbor representation in the embedding space obtained from vanilla neural networks. Let $\mathbf{H}$ represent the hidden feature learned by a standard neural network, APPNP aims at learning a context-aware representation $\mathbf{Z}$ as follows:

\begin{equation}
\mathbf{Z}^{(k)}=(1-\alpha)\widetilde{\mathbf{A}} \mathbf{Z}^{(k-1)} + \alpha \mathbf{H},
\end{equation}
where $\alpha \in [0,1]$ controls trade-off between the self-representation and contextual neighbor information. This update scheme has a tight relationship with diffusion processes. In fact, it is derived from the personalized PageRank retrieval system \cite{page1999pagerank}.

There are many other GCN models, such as GraphSAGE \cite{hamilton2017inductive} which proposes the idea of neighbor sampling,  
JK-nets \cite{xu2018representation}, which introduces residual connections to GCN. We refer readers to \cite{zhou2018graph,wu2019comprehensive} for comprehensive reviews.

\section{Manifold Smoothness Loss}
The central problem of GCNs is how to encode the graph structure, which is high-dimensional and non-Euclidean information, into feature vectors. Existing GCNs generally make use of the graph structure in every graph convolutional layers, where a graph-dependent neighbor aggregation scheme is integrated with layer-wise operations of vanilla neural networks. A typical GCN is built by stacking multiple these layers, and the model is trained by backpropagation with a specific loss function of choice. We focus on semi-supervised classification, as did by most GCNs, in which the cross-entropy loss $\mathcal{L}_{CE}$ is usually used to measure the difference between the predicted label distribution $\mathbf{z}_i$ and the graph-truth one-hot-encoded label indicator $\mathbf{y}_i$ as:

\begin{equation}
\mathcal{L}_{CE} = -\sum_{i \in L} \mathbf{y}_i \log{\mathbf{z}_i},
\end{equation}
where $L$ is the set of labeled points. This is problematic as there are limited labeled points in semi-supervised scenarios. GCNs trained by gradient descent with only the cross-entropy loss will quickly reach to perfect training accuracy and zero gradients, resulting in overfitting. Instead, we propose a loss function that consists of two terms:

\begin{align}\label{ch7:eq:loss_function}
&\mathcal{L} = \mathcal{L}_{fit} + \mu \, \mathcal{L}_{smooth},  \nonumber \\
&\text{s.t.} \quad \mathcal{L}_{fit} = \sum_{i \in L} f(\mathbf{z}_i, \, \mathbf{y}_i), \quad \mathcal{L}_{smooth} = \sum_{i,j \in L\cup U} A_{ij} \, f(\mathbf{z}_i, \, \mathbf{z}_j),
\end{align}
where $\mu \in [0,1]$ is a trade-off parameter, $U$ is the set of unlabeled points, $A_{ij}$ is an element of the adjacency matrix $\mathbf{A}$, and $f$ is a specific loss function of choice. While $\mathcal{L}_{fit}$ is a supervised loss that measures the fitness of the prediction $\mathbf{z}_i$ to the ground-truth $\mathbf{y}_i$ as other GCNs, the key point is the unsupervised loss $\mathcal{L}_{smooth}$ that imposes a regularization on the prediction itself. It favors a prediction function that is sufficiently smooth with respect to the graph structure, \ie, neighbor nodes on the graph should have similar predictions.

\subsection{The Choice of $\mathcal{L}_{fit}$, $\mathcal{L}_{smooth}$ and its Justification}
The loss functions $\mathcal{L}_{fit}$ and $\mathcal{L}_{smooth}$ measures the difference between two label vectors. They can be any loss functions, such as mean square error, cross-entropy loss, KL-divergence, and they are not necessarily to be the same. We present two special choices here, in which $\mathcal{L}_{fit}$ and $\mathcal{L}_{smooth}$ are both squared loss or cross-entropy loss, and we show that they have a tight relation with the diffusion process used in this thesis and some other existing GCNs.

\subsubsection{$L_2$ Loss}
Let the proposed loss $\mathcal{L}$ take the following form, where both $\mathcal{L}_{fit}$ and $\mathcal{L}_{smooth}$ are the squared loss:

\begin{equation}
\mathcal{L} = \sum_{i \in L} \left(\mathbf{z}_i - \mathbf{y}_i\right)^2 + \mu \sum_{i,j \in L\cup U} A_{ij} \, \left(\mathbf{z}_i - \mathbf{z}_j\right)^2.
\end{equation}
This is closely related to the graph regularization framework proposed in \cite{zhou2004learning}. The minimal loss can be obtained by setting the derivative with respect to $\mathbf{z}$ to 0, resulting in the solution of the prediction $\mathbf{z}$:

\begin{equation}\label{ch7:eq:z_solution}
\mathbf{z}^\ast = \gamma \big(\mathbf{I} - (1-\gamma)\mathbf{A})\big)^{-1}\mathbf{y},
\end{equation}
where $\gamma = \frac{1}{\mu+1}$. Indeed, $(\mathbf{I} - (1-\gamma)\mathbf{A})\big)^{-1}$ is a diffusion kernel, and it was used by another GCN~\cite{jiang2019data} in the layer-wise propagation for feature diffusion, except that they used $\mathbf{D}^{-\frac{1}{2}}\mathbf{A}\mathbf{D}^{-\frac{1}{2}}$ instead of $\mathbf{A}$. In this sense, the proposed loss function can be viewed as a label diffusion method that propagates labels from the labeled points to unlabeled points on the graph using a diffusion kernel.

Thanks to the connection between graph regularization framework and iterative diffusion process \cite{zhou2004learning, bai2017regularized}, it is easy to derive an iterative formula that converges to the same closed-form solution as in Eq.(\ref{ch7:eq:z_solution}), and it can be written in matrix form:

\begin{equation}
\mathbf{Z}^{(k)} = (1-\gamma)\mathbf{A}\mathbf{Z}^{(k-1)} + \gamma\mathbf{Y},
\end{equation}
where $\mathbf{Z}^{(0)}=\mathbf{Y}$, though the convergence does not depend on the initial condition. One may notice that this brings us to the update scheme of one of the state-of-the-art GCNs, APPNP~\cite{klicpera2018predict}. The only difference is that they used this iterative process to aggregate the neighbor's representation. While we directly aggregate neighbor's prediction. APPNP motivated their formula to the PageRank retrieval system. The proposed loss function, from this perspective, can be viewed as aggregating neighbors' prediction (the first term), with a certain chance of re-initialization (the second term). Optimizing the loss function is equivalent to run the iterative process infinity times until convergence, which significantly improves the model capacity on capturing long-range neighbor relationships. The existence of the re-initialization is then useful for preventing the over-smoothing effect \cite{li2018deeper}.

\subsubsection{Cross-Entropy Loss}
Another possible choice is the cross-entropy loss, which is proven to be good for classification. Instead of considering the total loss on all classes, the cross-entropy loss concentrates on the loss of the true class and imposes a large penalty when the predicted probability of belonging to the true class is low. In the proposed loss function Eq.(\ref{ch7:eq:loss_function}), it is straightforward to use cross-entropy loss for $\mathcal{L}_{fit}$. However, directly applying it for $\mathcal{L}_{smooth}$ is problematic as both predictions $\mathbf{z}_i$ and $\mathbf{z}_j$ are not one-hot-encoded vectors. Cross-entropy on these vectors will lose the focus on the true class and lead to poor classification performance. Therefore, we convert one of the predictions to a one-hot-encoded vector before we apply the cross-entropy loss. The conversion can be represented by the following function:

\begin{equation}
\phi{(\mathbf{z}_{ij})} = 
\begin{cases}
1, \quad \text{if} \,\, \mathbf{z}_{ij} = \max(\mathbf{z}_i), \\
0, \quad \text{otherwise},
\end{cases}
\end{equation}
where $\phi$ is an element-wise function. Putting $\mathcal{L}_{fit}$ and $\mathcal{L}_{smooth}$ together, we can write the proposed loss function as:

\begin{equation}
\mathcal{L} = -\sum_{i \in L} \mathbf{y}_i \log{\mathbf{z}_i} - \mu \sum_{i \in L\cup U} A_{ij} \, \phi(\mathbf{z}_i) \log{\mathbf{z}_j}.
\end{equation}
The first term is the supervised cross-entropy loss for the labeled points used by most of the existing GCNs, while the second term is an unsupervised entropy-loss that works on both the labeled and unlabeled points. The unsupervised cross-entropy loss enforces the neighbor's predictions to have the maximal value for the same class, \ie, it enforces neighbors to be the same label. Note that the loss on the labeled points has been measured in both terms, but they are not redundant as one measures the fitness to the ground-truth, and another regularizes the label smoothness on the graph.

\section{Experiments}
To demonstrate the effectiveness of the proposed smoothness loss, we conduct various controlled experiments on both MLP and state-of-the-art GCNs. They are evaluated with and without the smoothness loss, as we concern about the relative difference rather than the absolute performance. Following most GCNs, we focus on the task of semi-supervised classification of citation networks, and we show that the proposed loss can not only significantly improve vanilla MLP but also bring performance gain for all the state-of-the-art GCNs.

\subsection{Experimental Settings}
We evaluated the proposed loss on MLP, and several popular GCNs, including GCN \cite{kipf2017semi}, GAT \cite{velivckovic2017graph}, and APPNP \cite{klicpera2018predict}. As the proposed smoothness loss can be essentially viewed as a regularization, we refer these methods used with the proposed loss as R-MLP, R-GCN, R-GAT, R-APPNP for the sake of simplicity and easy comparison. For MLP, we use the vanilla MLP without any bells and whistles, and each layer is a fully-connected layer with 64 hidden units. For GCNs, we use the same set of hyper-parameters before and after adding the smoothness loss for fair comparisons. Specifically, we use all GCNs with two hidden layers, 64 hidden units, ReLU activation, dropout layers before each hidden layer at rate 0.5, $\ell_2$ regularization at $5e^{-4}$, and learning rate 0.01. For the smoothness loss, we use the modified cross-entropy loss as it is better for classification tasks. All the weights are initialized by the method described in \cite{glorot2010understanding}, and we train all GCNs for a maximum of 1000 epochs using Adam \cite{kingma2014adam} and early stopping with a window size 100, \ie, we stop training if the validation loss does not decrease for 100 consecutive epochs.

We conduct experiments on the wide-used citation datasets, including Cora, Citeseer, and Pubmed \cite{sen2008collective}, the information of interest is summarized in Table \ref{ch7:tab:citation_datasets}. These datasets consist of documents and citation links between them. Documents are represented by sparse bag-of-words feature vectors, and the citation links are treated as undirected edges used for constructing the binary adjacency matrix $\mathbf{A}$. For all the citation datasets, we random split 50 times, and the average performance is presented. Each split contains $\ell$ nodes per class, 500 validation nodes, and 1000 testing nodes.

\begin{table}[htp]
	\caption{Summary of datasets used in the experiments.}
	\label{ch7:tab:citation_datasets}
	\centering
	\resizebox{0.9\linewidth}{!}{
		\begin{tabular}{ccccccc}
			\toprule
			\textbf{Dataset} & \textbf{Nodes} & \textbf{Edges} & \textbf{Classes} & \textbf{Features} & \textbf{Validation Nodes} & \textbf{Testing Nodes} \\
			\midrule
			Cora & 2,708 & 5,429 & 7 & 1,433 & 500 & 1,000 \\
			Citeseer & 3,327 & 4,732 & 6 & 3,703 & 500 & 1,000 \\
			Pubmed & 19,717 & 44,338 & 3 & 500 & 500 & 1,000\\
			\bottomrule
		\end{tabular}
	}
\end{table}

\subsection{Qualitative Comparison of Embedding Visualization}
The core idea of GCNs is to make use of graph structure to learn a good representation that can be used for downstream machine learning tasks, such as classification. To this end, existing GCNs usually integrate an aggregation scheme to the layer-wise operations of neural networks so as to explicitly learn context-aware representation. In contrast, we propose to regularize the loss function with a graph smoothness loss defined on both labeled points and unlabeled points. Despite that, we do not model node representation explicitly, GCNs used in conjunction with the proposed loss tend to produce more compact representations that are better for discriminative analysis. As we push neighbor node's predictions to be close, there is an implicit effect to enforce the neighbor node's representations to be close as well, which results in compact classes that are ready to be classified.

\begin{figure}[htb]
\begin{center}
  \includegraphics[width=0.9\linewidth]{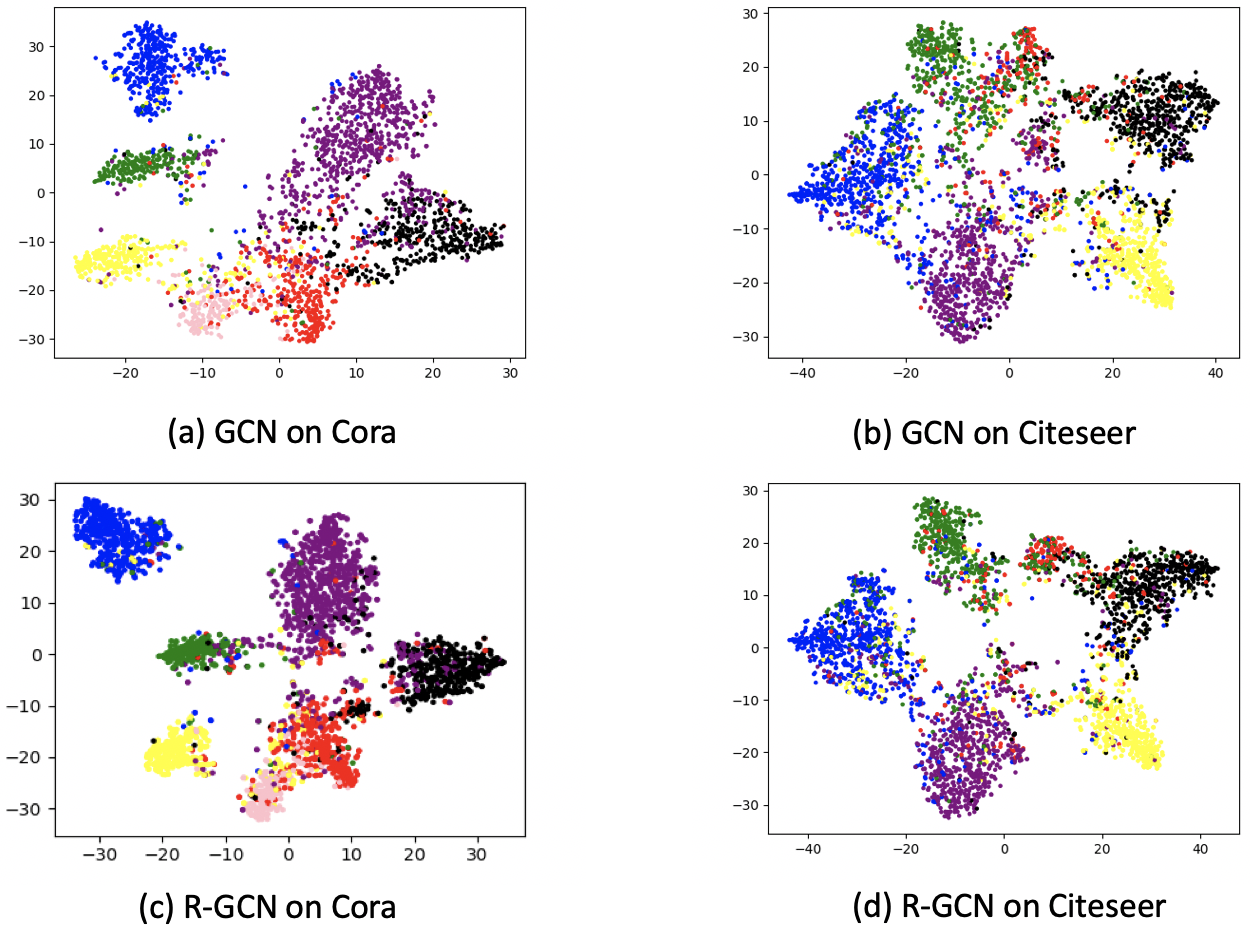}
\end{center}
  \caption{T-sne plots of the learned representations of GCN and R-GCN on Cora and Citeseer datasets.}
  \label{ch7:fig:tsne_plots}
\end{figure}

We train a two-layer GCN model with and without the regularization loss on the Cora and Citeseer datasets, denoted by GCN and R-GCN. We extract the output of the last hidden layer as the learned representation (embedding) and visualize it in two-dimensional space using t-sne \cite{maaten2008visualizing}. As shown in Fig.\ref{ch7:fig:tsne_plots}, the representations obtained by R-GCN are clearly more compact than those of GCN, which demonstrates the effectiveness of learning useful representation of the proposed smoothness loss. This also explains why it can improve classification accuracy.

\subsection{Quantitative Comparison of Classification Accuracy}
\subsubsection{Baseline Multi-Layer Perceptrons}
Next, we conduct quantitative comparisons between models and their regularized counterparts on semi-supervised classification. We hypothesized that limited labeled points could not train a deep model, and proposed an unsupervised loss to address it. To demonstrate this, we first compare the vanilla MLP with R-MLP using a different number of hidden layers. The classification accuracy on the Cora, Citeseer, Pubmed datasets are reported.

As shown in Table \ref{ch7:tab:mlp_classification}, the performance of vanilla MLP is significantly improved by adding the proposed smoothness loss, which also shows the importance of making use of graph structure when it is available. On the other hand, the performance of R-MLP also demonstrates the potential capability of the smoothness loss on training deep architectures. 
While vanilla MLP reaches its optimal performance with 1 or 2 layers and drops dramatically, adding the smoothness loss makes its performance much more stable and performs reasonably even with five hidden layers. 

\begin{table}[htp]
	\caption{Semi-Supervised Classification Accuracy $\pm$ Standard Deviation (\%) of MLPs.}
	\label{ch7:tab:mlp_classification}
	\centering
	\resizebox{1\linewidth}{!}{
		\begin{tabular}{c|cccccc}
			\toprule
			\textbf{Datasets} & \textbf{\# Layers} & 1 & 2 & 3 & 4 & 5 \\
			\midrule
			\multirow{2}{*}{Cora} & MLP  & 57.5 $\pm$ 2.4 & 57.7 $\pm$ 2.0 & 54.3 $\pm$ 3.5 & 33.2 $\pm$ 8.2 & 26.6 $\pm$ 7.3 \\
								 & R-MLP & \textbf{58.6 $\pm$ 2.1} & \textbf{76.0 $\pm$ 2.3} & \textbf{77.4 $\pm$ 3.1} & \textbf{77.6$\pm$ 4.2} & \textbf{73.3$\pm$ 4.7} \\
			\midrule
			\multirow{2}{*}{Citeseer} & MLP  & 58.8 $\pm$ 2.3 & 57.0 $\pm$ 2.0 & 52.1 $\pm$ 2.9 & 36.5 $\pm$ 7.9 & 26.5 $\pm$ 5.6 \\
								   & R-MLP & \textbf{60.6 $\pm$ 2.1} & \textbf{62.5 $\pm$ 2.3} & \textbf{62.7 $\pm$ 2.5} & \textbf{57.4 $\pm$ 2.8} & \textbf{52.2 $\pm$ 4.8} \\			
			\midrule
			\multirow{2}{*}{Pubmed} & MLP  & 69.9 $\pm$ 2.7 & 69.6 $\pm$ 2.5 & 68.0 $\pm$ 2.5 & 65.9 $\pm$ 5.3 & 56.7 $\pm$ 7.9 \\
								  & R-MLP & \textbf{70.6 $\pm$ 2.4} & \textbf{70.7$\pm$ 2.7} & \textbf{69.3$\pm$ 3.2} & \textbf{66.5$\pm$ 5.1} & \textbf{61.1$\pm$ 6.5} \\			
			\bottomrule
		\end{tabular}
	}
\end{table}

\subsubsection{State-of-the-art Graph Convolutional Netowrks}
Last but not least, we show that the proposed smoothness is ready to be plugged into any state-of-the-art GCN models to improve its performance. We compare the classification performance of GCN, GAT, APPNP to their regularized version in the semi-supervised setting. In addition, to study the effect of the number of labeled points on the model accuracy, we use a different number of labeled points per class, as denoted by $\ell$. $\ell$ is tested for 20, 10, 5, and the classification accuracy is reported, respectively.

\begin{table}[htp]
	\caption{Semi-Supervised Classification Accuracy $\pm$ Standard Deviation (\%) of GCNs.}
	\label{ch7:tab:gcn_classification}
	\centering
	\resizebox{1\linewidth}{!}{
		\begin{tabular}{c|c|cc|cc|cc}
			\toprule
			\textbf{Datasets} & \textbf{$\ell$} & GCN & R-GCN & GAT & R-GAT & APPNP & R-APPNP \\
			\midrule
			\multirow{3}{*}{Cora} & 20  & 79.2 $\pm$ 1.7 & \textbf{81.9} $\pm$ \textbf{1.8} & 80.8 $\pm$ 1.6 & \textbf{81.9} $\pm$ \textbf{1.3} & 82.7 $\pm$ 1.3 & \textbf{83.4} $\pm$ \textbf{1.4} \\
								 & 10  & 75.8 $\pm$ 2.2 & \textbf{78.6} $\pm$ \textbf{2.2} & 77.6 $\pm$ 2.2 & \textbf{78.2} $\pm$ \textbf{2.1} & 80.1 $\pm$ 1.5 & \textbf{81.2} $\pm$ \textbf{2.1} \\
								 & 5  & 68.1 $\pm$ 4.1 & \textbf{71.9} $\pm$ \textbf{4.2} & 71.8 $\pm$ 3.4 & \textbf{72.3} $\pm$ \textbf{3.5} & 75.1 $\pm$ 3.5 & \textbf{77.2} $\pm$ \textbf{2.9} \\
			\midrule
			\multirow{2}{*}{Citeseer} & 20  & 68.3 $\pm$ 2.0 & \textbf{70.9} $\pm$ \textbf{2.3} & 68.5 $\pm$ 1.6 & \textbf{69.4} $\pm$ \textbf{1.4} & 69.9 $\pm$ 1.6 & \textbf{71.4} $\pm$ \textbf{1.5} \\
								 	 & 10  & 64.8 $\pm$ 2.6 & \textbf{67.3} $\pm$ \textbf{2.7} & 66.2 $\pm$ 2.5 & \textbf{67.1} $\pm$ \textbf{2.2} & 67.0 $\pm$ 2.7 & \textbf{68.6} $\pm$ \textbf{2.6} \\
								   & 5  & 57.4 $\pm$ 4.2 & \textbf{58.7} $\pm$ \textbf{3.6} & 59.4 $\pm$ 4.6 & \textbf{61.4} $\pm$ \textbf{4.3} & 61.6 $\pm$ 3.8 & \textbf{62.8} $\pm$ \textbf{3.6} \\
			\midrule
			\multirow{2}{*}{Pubmed} & 20  & 77.5 $\pm$ 2.5 & \textbf{78.6} $\pm$ \textbf{2.5} & 77.5 $\pm$ 2.5 & \textbf{78.3} $\pm$ \textbf{2.0} & 79.1 $\pm$ 2.5 & \textbf{80.9} $\pm$ \textbf{2.2} \\
								 	& 10  & 73.8 $\pm$ 3.6 & \textbf{75.1} $\pm$ \textbf{3.4} & 73.7 $\pm$ 4.1 & \textbf{75.3} $\pm$ \textbf{2.5} & 76.2 $\pm$ 3.4 & \textbf{77.8} $\pm$ \textbf{2.9} \\
								  & 5  & 68.9 $\pm$ 4.5 & \textbf{69.7} $\pm$ \textbf{4.5} & 67.8 $\pm$ 5.0 & \textbf{70.3} $\pm$ \textbf{4.8} & 72.2 $\pm$ 5.5 & \textbf{73.4} $\pm$ \textbf{4.5} \\			
			\bottomrule
		\end{tabular}
	}
\end{table}

As shown in Table \ref{ch7:tab:gcn_classification}, adding the smoothness loss improves the classification accuracy of all GCN models consistently, including GCN, GAT, and APPNP. This seems counter-intuitive, as the graph structure has been used by these GCNs already in the layer-wise propagations, which means using it again in the loss function seems redundant. We hypothesize that the graph information has not been extensively exploited by existing GCNs due to their inability to capturing long-range neighbor relationships. Coupling of neighbor aggregation and the layer-wise operation takes existing GCNs to a dilemma between strict localization (shadow model) and over-smoothing (deep model), while the proposed smoothness loss can model global graph structure without going deep. On the other hand, different values of $\ell$ demonstrate the proposed smoothness loss is robust with respect to the number of labeled points, as it can still improve the performance even with only five labeled points per class.

\section{Summary}
In this work, we discuss the representation learning on graphs with graph convolutional networks (GCNs). While most existing GCNs focus on encoding graph structure into layer-wise neural operations by a neighbor aggregation scheme, we propose a simple yet effective manifold smoothness loss to regularize these GCNs. It addresses two problems of existing GCNs, which are overfitting and local-only. We justify the proposed loss by drawing its connection to other GCNs and the diffusion process and show that minimizing the proposed loss can be viewed as aggregating neighbor predictions with infinity layers. The overfitting problem is alleviated as the smoothness loss is unsupervised so that we can take advantage of the unlabeled points. The long-range neighbor relationships are captured without adding more layers explicitly, which reduces the risk of over-smoothing. 

We conduct qualitative comparisons to show that GCNs trained in conjunction with the smoothness loss learn more compact representation than their vanilla counterparts. Some quantitative analyses demonstrate that such representations can yield better performances on downstream discriminative machine learning tasks, such as semi-supervised classification.

\subsubsection*{Acknowledgments}
This work is supported by a Research Training Program scholarship of Australia government.

\bibliography{rgcn}
\bibliographystyle{apalike}

\end{document}